\documentclass[runningheads]{llncs}
\usepackage{graphicx}
\usepackage{amsmath,amssymb} 
\usepackage{color}

\usepackage{makeidx}
\usepackage{graphicx}
\usepackage{amsmath,amssymb} 
\usepackage{amssymb,mathtools}
\usepackage{color}

\usepackage[printonlyused,nolist]{acronym}
\usepackage{pgfplots}

\usepackage{soul}
\usepackage{makeidx}
\usepackage{graphicx}
\usepackage{amsmath,amssymb} 
\usepackage{color}
\usepackage{subfig}
\usepackage{caption}
\usepackage{rotating}
\usepackage{array}
\usepackage{todonotes}
\usepackage{color, colortbl}
\usepackage{collcell}
\usepackage{hhline}
\usepackage{pgf}
\pgfplotsset{compat=1.9}

\usepackage{enumitem}
\usepackage{algorithm}
\usepackage{algorithmic}

\newcommand {\etal}{\emph{et al.}}
\newcommand {\eg}{\emph{e.g.}}
\newcommand {\ie}{\emph{i.e.}}
\newcommand {\figref}[1]{Fig.~\ref{fig:#1}}
\newcommand {\equref}[1]{Eq.~\ref{eq:#1}}
\newcommand {\tableref}[1]{Table.~\ref{table:#1}}

\begin{document}

\begin{acronym}
\acro{CNN}{Convolutional Neural Network}
\acro{SCE}{Softmax Cross Entropy}
\acro{CE}{Cross Entropy}
\acro{TCE}{Tamed Cross Entropy}
\acro{MSE}{Mean Squared Error}
\acro{MAE}{Mean Absolute Error}
\acro{NLL}{Negative Log Likelihood}
\acro{TNLL}{Tamed Negative Log Likelihood}
\end{acronym}

\title{Taming the Cross Entropy Loss}

\titlerunning{Taming the Cross Entropy Loss}
\author{Manuel Martinez \and Rainer Stiefelhagen}
\authorrunning{M. Martinez, R. Stiefelhagen}
\institute{Kalrsruhe Institute of Technology, Karlsruhe, Germany\\
\email{\{manuel.martinez, rainer.stiefelhagen\}@kit.edu}
}

\maketitle              

\begin{abstract}
We present the Tamed Cross Entropy (TCE) loss function, a robust derivative of the standard Cross Entropy (CE) loss used in deep learning for classification tasks.
However, unlike other robust losses, the TCE loss is designed to exhibit the same training properties than the CE loss in noiseless scenarios.
Therefore, the TCE loss requires no modification on the training regime compared to the CE loss and, in consequence, can be applied in all applications where the CE loss is currently used.
We evaluate the TCE loss using the ResNet architecture on four image datasets that we artificially contaminated with various levels of label noise.
The TCE loss outperforms the CE loss in every tested scenario.
\end{abstract}

\section{Introduction}
The most common way to train \acp{CNN} for classification problems is to use stochastic gradient descent coupled with the \acf{CE} loss.
The \ac{CE} loss is popular mainly due to its excellent convergence speeds, alongside its excellent performance in terms of Top-1 and Top-5 classification accuracy.

However, the \ac{CE} loss is not without weaknesses.
Theoretically, the \ac{CE} is proven to be a calibrated loss~\cite{tewari2007consistency}, and thus should provide well-behaved probability estimates,
however, in reality a different behavior is observed: the calibration of a classifier using the \ac{CE} loss worsens as the classification accuracy improves~\cite{guo2017calibration}.
As a consequence, many techniques have have been proposed to improve calibration (\eg, Bayesian Neural Networks~\cite{cobb2018loss}).


A related problem of the \ac{CE} loss is its suboptimal performance when dealing with noisy data~\cite{ghosh2017robust}.
Although complex \acp{CNN} architectures have shown considerable robustness to noise in the training dataset~\cite{flatow2017robustness,prakash2018protecting,rolnick2017deep}, noisy labels and outliers are still a significant problem, particularly when dealing with weak labels.
As a consequence, the problem of dealing with label noise when learning has been studied extensively~\cite{frenay2014classification}.

In particular, there are loss functions for classification tasks that are more robust or have more discriminative power than the \ac{CE}.
For example, the pairwise loss~\cite{hadsell2006dimensionality} and the triplet loss~\cite{schroff2015facenet} are effective ways to learn discriminative features between individual classes.
Also, the OLE loss~\cite{lezama2018ole} explicitly maximizes intra-class similarity and inter-class margin, and thus, improves its discriminative power with respect to the \ac{CE}.
However, such losses are either slower or significantly more complex to apply than the \ac{CE}.

Ghosh \etal~\cite{ghosh2017robust} used a risk minimization framework to analyze the \ac{CE} loss, the \ac{MAE} loss, and the \ac{MSE} loss, for classification tasks under artificially added label noise.
Their results show that the \ac{MAE} is inherently robust to noise, while the \ac{CE} is particularly vulnerable to label noise, and the \ac{MSE} should perform better than the \ac{CE} but worse than the \ac{MAE}.
Sadly, being an $\ell_1$ loss, the \ac{MAE} has abysmal convergence properties and is not well suited for practical use.

We aim to offer a more convenient alternative to the currently available losses for robust classification.
We follow the same spirit than Huber \etal~\cite{huber1964robust} and Girshick \etal~\cite{girshick2015fast}, who independently hand crafted a robust regression loss by fusing the \ac{MSE} loss and the \ac{MAE} loss together, and thus obtained a loss with the convergence properties of the \ac{MSE}, and the robustness to noise of the \ac{MAE}.

Our result is the \acf{TCE} loss, which is derived from the \ac{CE} and thus it shares the same convergence properties, while, at the same time, its more robust to noise.
Instead of fusing two losses, we started from the \ac{CE} and designed guidelines on how the gradient of our tentative \ac{TCE} should behave in order to behave like the \ac{CE} and be robust to outliers.

Finally, to design the actual \ac{TCE}, we used a power normalization over the \ac{CE} gradient to make it compatible with our previously designed guidelines.
We choose this kind of regularization because power normalizations have already been used with great success to robustify features~\cite{koniusz2013comparison}.

The gradient of the \ac{TCE} is identical to the gradient of the \ac{CE} if the predicted confidence with respect to the actual label is high, and tends to zero if the predicted confidence of with respect to the actual label is low. 
This way, training samples that produce low confidence values (ideally outliers or misslabeled data), generate a reduced feedback response.

To ensure that the \ac{TCE} can be used as a drop-in replacement for the \ac{CE}, we used the reference implementation for the ResNet~\cite{he2016deep} architecture and we replaced the \ac{CE} with the \ac{TCE} without altering any configuration parameters.
We also tested the performance of the \ac{TCE} against the \ac{CE}, the \ac{MSE}, and the \ac{MAE} losses in the same scenario, and we also evaluated the robustness of the loss functions against uniformly distributed label noise.

In all tested cases, our \ac{TCE} outperformed the \ac{CE} while having almost the same convergence speed.
Furthermore, with 80\% of random labels, the \ac{TCE} offers Top-1 accuracy improvements of $9.36\%$ , $9.80\%$ , and $4.94\%$ in CIFAR10+, CIFAR100+, and VSHN respectively.

\clearpage

\section{Taming the \acl{NLL} Loss}

\subsection{Background}

The cross entropy loss is commonly used after a softmax layer that normalizes the output of the network, and is defined as:

\begin{equation}
\label{eq:softmax}
\text{Softmax}(\mathbf{o}) = \frac{e^\mathbf{o}}{\sum_{j=1}^N e^{\mathbf{o}_j}}\,,
\end{equation}
whereas the cross entropy between two $N$ sized discrete distributions $\mathbf{p} \in [0,1]^N$ and $\mathbf{q} \in (0,1]^N$ is:
\begin{equation}
\label{eq:ce}
H(\mathbf{p}, \mathbf{q}) = -\sum^N_{i=1} \mathbf{p}_i\, \log \mathbf{q}_i \,,
\end{equation}
where $\mathbf{p}$ corresponds to the classification target, and $\mathbf{q}$ the output of the softmax layer, \ie, the likelihood predicted per class.

Is it important to note that the actual value of the loss function does not affect in any way the training procedure, as only its gradient is used during back propagation.
We analyze the gradient of the cross entropy loss with respect to the log-likelihood, which is a commonly used trick.
Using $\mathbf{p} \in \{0,1\}^N$, the partial derivatives of the \ac{CE} loss with respect to the predicted log-likelihoods are:

\begin{equation}
\label{eq:nllg}
\frac{\partial H(\mathbf{p}, \mathbf{q})}{\partial \log \mathbf{q}_i} =
  \begin{cases}
    0 & \text{if $\mathbf{p}_i = 0$,} \\
    -1 & \text{if $\mathbf{p}_i = 1$.}
  \end{cases}
\end{equation}

\subsection{Design Goals}

We define the following set of design goals in order to guide us in the design process towards a robust classification goal:





\begin{enumerate}
\item We want the gradient of the \ac{TCE} loss ($\hat{H}$) to be proportional to the gradient of the \ac{CE} loss. 
This way we expect that both losses will behave in a similar way.
We aim to:
\begin{equation}
\label{eq:cestg}
\nabla \hat{H}(\mathbf{p}, \mathbf{q}) \propto \nabla H(\mathbf{p}, \mathbf{q}) \, .
\end{equation}

\item If the network is confident about the predicted class, \ie, $\mathbf{q}_i \to 1$, we want the \ac{TCE} to behave exactly like the \ac{CE}.
\begin{equation}
\label{eq:cestg}
\nabla \hat{H}(\mathbf{p}, \mathbf{q}) = -1 \quad \text{if $p_i = 1$ and $q_i \to 1$} \, .
\end{equation}

\item We aim to reduce the impact of outliers by reducing the feedback from the gradient when there is a large discrepancy between a prediction and its associated label:
\begin{equation}
\label{eq:cestg}
\nabla \hat{H}(\mathbf{p}, \mathbf{q}) = 0 \quad \text{if $p_i = 1$ and $q_i \to 0$} \, .
\end{equation}

\end{enumerate}

To summarize, we aim to design a function whose gradient behaves in the following way:

\begin{equation}
\label{eq:goal}
\frac{\partial \hat{H}(\mathbf{p}, \mathbf{q})}{\partial \log \mathbf{q}_i} \approx
  \begin{cases}
    0 & \text{if $\mathbf{p}_i = 0$  ,} \\
    -1 & \text{if $\mathbf{p}_i = 1$ and $\mathbf{q}_i \to 1$ ,} \\
    0 & \text{if $\mathbf{p}_i = 1$ and $\mathbf{q}_i \to 0$ .}
  \end{cases}
\end{equation}

\subsection{The gradient of the \acl{TCE} Loss}

We suggest the following gradient that fulfills the requirements expressed in \equref{goal}:
\begin{equation}
\label{eq:deriv}
\frac{\partial \hat{H}_\alpha(\mathbf{p}, \mathbf{q})}{\partial \log \mathbf{q}_i} =
  \begin{cases}
    0 & \text{if $\mathbf{p}_i = 0$ ,} \\
    -(1 - \log \mathbf{q}_i)^{-\alpha} & \text{if $\mathbf{p}_i = 1$ .}
  \end{cases}
\end{equation}
We based our regularization on the domain $[1,\infty)$ of the power function, which we applied to the $\log \mathbf{p}_i$ term.
And we control the regularization factor using the parameter $\alpha \in \mathbb{R}^+$.

The loss function that corresponds with the gradient presented in \equref{deriv} is:

\begin{equation}
\label{eq:ce}
\hat{H}_\alpha(\mathbf{p}, \mathbf{q}) = \frac{1}{1-\alpha} \sum^N_{i=1} \mathbf{p}_i\, \left( \left( 1 - \log \mathbf{q}_i \right) ^{1-\alpha} - \frac{1}{1-\alpha} \right) \,.
\end{equation}

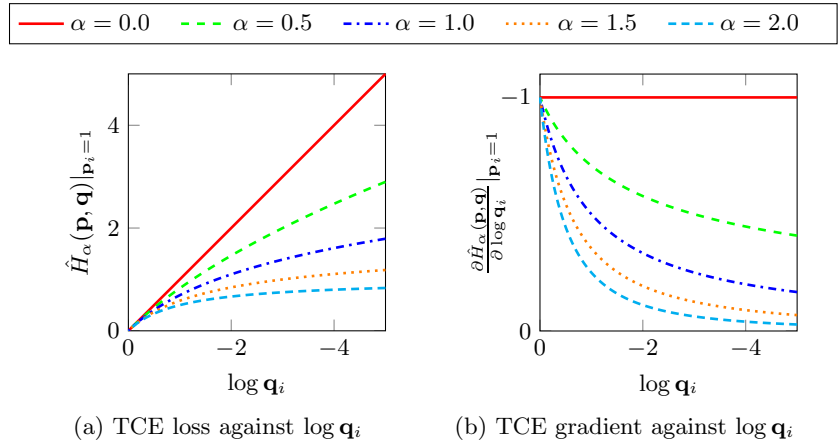
\begin{figure}[t]
\centering
\begin{tikzpicture}
    \begin{axis}[%
    hide axis,xmin=10,xmax=50,ymin=0,ymax=0.4,
	legend columns=5,legend style={cells={align=left}}
    ]
    \addlegendimage{line width=1pt, solid, color=red}
    \addlegendentry{$\alpha = 0.0 \quad$};
    \addlegendimage{line width=1pt, dashed, color=green}
    \addlegendentry{$\alpha = 0.5 \quad$};
    \addlegendimage{line width=1pt, dash dot, color=blue}
    \addlegendentry{$\alpha = 1.0 \quad$};
    \addlegendimage{line width=1pt, dotted, color=orange}
    \addlegendentry{$\alpha = 1.5 \quad$};
    \addlegendimage{line width=1pt, densely dashed, color=cyan}
    \addlegendentry{$\alpha = 2.0 \quad$};
    \end{axis}
\end{tikzpicture}

\subfloat[\ac{TCE} loss against $\log \mathbf{q}_i$]{
\begin{tikzpicture}
    \begin{axis}[
		height=5cm,  width=5cm,
		samples=128,xmin=-5,xmax=0,ymin=0, ymax=5,domain=-5:0,x dir=reverse,
		y label style={at={(axis description cs:-0.1,.5)},rotate=0,anchor=south},
		xlabel={$\log \mathbf{q}_i$},
		ylabel={$\hat{H}_\alpha(\mathbf{p}, \mathbf{q}) |_{\mathbf{p}_i=1}$}]
    \addplot[line width=1pt, solid, color=red] {pow(abs(1-x),1)-1}; 
    \addplot[line width=1pt, dashed, color=green] {(1/0.5)*pow(abs(1-x),0.5)-(1/0.5)}; 
    \addplot[line width=1pt, dash dot, color=blue] {ln(1-x)};
    \addplot[line width=1pt, dotted, color=orange] {(1/-0.5)*pow(abs(1-x),-0.5)+2};
    \addplot[line width=1pt, densely dashed, color=cyan] {(1/-1)*pow(abs(1-x),-1)-(1/-1)};
    \end{axis}
\end{tikzpicture}}
~~~~~~~
\subfloat[\ac{TCE} gradient against $\log \mathbf{q}_i$]{
\begin{tikzpicture}
    \begin{axis}[
		height=5cm,  width=5cm,
		ytick={0,-1},
		samples=128,xmin=-5,xmax=0,ymin=-1.1, ymax=0,domain=-5:0,x dir=reverse, y dir=reverse,
		y label style={at={(axis description cs:-0.1,.5)},rotate=0,anchor=south},
		xlabel={$\log \mathbf{q}_i$},
		ylabel={$\frac{\partial \hat{H}_\alpha(\mathbf{p}, \mathbf{q})}{\partial \log \mathbf{q}_i} |_{\mathbf{p}_i=1}$}]
    \addplot[line width=1pt, solid, color=red] {-pow(abs(1-x),1-1)}; 
    \addplot[line width=1pt, dashed, color=green] {(-pow(abs(1-x),.5-1)}; 
    \addplot[line width=1pt, dash dot, color=blue] {-pow(abs(1-x),0-1)};
    \addplot[line width=1pt, dotted, color=orange] {-pow(abs(1-x),-0.5-1)};
    \addplot[line width=1pt, densely dashed, color=cyan] {-pow(abs(1-x),-1-1)};
    \end{axis}
\end{tikzpicture}}
\caption{
The \ac{CE} (case $\alpha=0.0$) has a constant gradient when plotted against $\log \mathbf{q}_i$, thus its response is independent of the confidence estimate of the prediction. 
On the other hand, \ac{TCE}'s gradient gets smaller as the confidence estimate of the prediction decreases.
}
\label{fig:fandg}
\end{figure}

We can observe the behavior of both $\hat{H}_\alpha$ and $\nabla \hat{H}_\alpha$ in \figref{fandg}.
Also, note that $\hat{H}_\alpha$ corresponds to $H$, when $\alpha$ equals $0$.

\clearpage

\section{Experiments}

\subsection{Experimental Setup}

We evaluate the \ac{TCE} against the \ac{CE} loss and other baselines on four datasets: MNIST~\cite{lecun1998gradient}, CIFAR10~\cite{krizhevsky2009learning}, CIFAR100~\cite{krizhevsky2009learning}, and VSHN~\cite{netzer2011reading}.
All datasets are well known, and consist of 32x32 pixel images.
MNIST, CIFAR10, and VSHN contain 10 classes each, while CIFAR100 contains 100 different classes.

Our training setup is based on the reference implementation for the ResNet~\cite{he2016deep}, implemented in Torch~\cite{collobert2011torch7}.
We train the same architecture (ResNet-20) for all datasets, and we use the default training strategy, which is optimized for the \ac{CE} loss we aim to replace.
For CIFAR10 and CIFAR100 we apply common data augmentation schemes (shifting and mirroring), thus we decorate both datasets with a "+" mark on the evaluation.
On MNIST and VSHN we apply only shifting, as they depict numbers.
We normalize the data using the channel means and standard deviation.
We use an initial learning rate of $0.1$, a Nesterov momentum of $0.9$, batch size of $128$, and weight decay of $1e-4$.
We train for 256 epochs, and we divide the learning rate by 10 at epoch 128, and again at epoch 192.

We hold out 5000 images from the training set of each dataset and we use them as a validation set.
Such validation set is used only to determine at which epoch the lowest validation error is obtained.
Then, we run 5 times each experiment (using the entire training data) and we report the mean and the standard deviation (when significant) of the test error captured at the epoch determined by the previous validation step.

All losses can be computed efficiently, hence there is no discernible difference in time when training using different loss functions.

\subsection{Baselines}

We compare our \ac{TCE} loss to the \ac{CE} loss we aim to replace, as well as the \ac{MSE} and the \ac{MAE} losses, both suggested by Ghosh \etal~\cite{ghosh2017robust} as robust alternatives to the \ac{CE} loss.
The Huber loss~\cite{huber1964robust}, also known as SmoothL1 loss~\cite{girshick2015fast}, is a well known loss used in robust regression, however there is no need to evaluate it as it is equivalent to the \ac{MSE} loss when applied to the $[0,1]$ domain used for classification.

As both \ac{MSE} and \ac{MAE} are losses designed for regression problems, we had to adapt them prior to use them for classification.
We followed the Torch~\cite{collobert2011torch7} guidelines: we prefixed them with a Softmax later, and we scaled their gradient by the number of classes in the output.

\clearpage

\subsection{Top-1 Accuracy under Uniform Label Noise}

A common experiment to evaluate robustness in deep learning is to perform an experiment where we apply uniformly distributed random labels to a portion of the training dataset~\cite{flatow2017robustness,ghosh2017robust,jindal2016learning,rolnick2017deep}.
In this setup, the noise ratio ($\eta$) determines the proportion of the training dataset corrupted with random labels, and we evaluated our losses on the four datasets using $\eta \in \{0.0, 0.2, 0.4, 0.6, 0.8\}$.

We group the full results of this experiment on the challenging CIFAR100+ dataset in \tableref{resultsNoise100}.
We group the results on the 10-class datasets in the \tableref{resultsNoise10}, where we only show the results for $\eta \in \{0.0, 0.4, 0.8\}$ for space reasons.

We observe that, when using default training regimes, the \ac{MAE} norm fails to converge, something expected from a pure $\ell_1$ norm loss.
Although in~\cite{ghosh2017robust} it is argued that the \ac{MSE} should be more robust to noise than the \ac{CE}, the improvement is small and only occurs on low noise factors (\ie, $\eta \le 0.4$).
In general terms, both \ac{CE} and \ac{MSE} losses obtain similar performance. 

On the other hand, the \ac{TCE} losses achieve the best Top-1 accuracy in all but one case, where it is second best after \ac{MSE}.
For $\eta = 0.8$, the \ac{TCE} improves Top-1 accuracy by $9.36\%$ in CIFAR10+, $9.80\%$ in CIFAR100+, and $4.94\%$ in VSHN.
Furthermore, the \ac{TCE} loss shows little sensitivity to its regularization parameter, offering solid performances for $\alpha \in \{0.5,1.0,1.5,2.0\}$.

\begin{table}[t!]
\centering
\footnotesize
\def\arraystretch{1.2}
\begin{tabular}{ l | c c c c c | }
  & \multicolumn{5}{c|}{CIFAR100+} \\
  noise ($\eta$) & 0.0 & 0.2 & 0.4 & 0.6 & 0.8 \\ \hline \hline
  CE  & 68.18 &61.16 &54.52 &44.08 &20.30 \\ \hline
  MSE & 67.78 &62.81 &55.98 &42.48 &15.41 \\ \hline
  MAE & 1.00 &1.00 &1.00 &1.00 &1.00 \\ \hline
  TCE $\alpha=0.5$ & 68.25 &63.58 &57.88 &48.10 &25.13 \\
  TCE $\alpha=1.0$ & 68.33 &64.11 &59.90 &51.59 &29.58 \\
  TCE $\alpha=1.5$ & \bf{68.45} &\bf{65.10} &61.07 &\bf{53.76} &\bf{30.10} \\
  TCE $\alpha=2.0$ & 66.81 &64.37 &\bf{61.51} &52.09 &18.75 \\ \hline
\end{tabular}

Top-1 Accuracy (\%)
\vspace{1em}
\caption{
Evaluation of our \ac{TCE} loss against alternatives using a ResNet-20 on CIFAR100+.
We used the reference ResNet implementation and default parameters except for the loss function.
A proportion of the training dataset ($\eta$) had its labels replaced randomly. 
CIFAR100+ is a complex dataset with 100 classes, and thus we can observe how neither \ac{MSE} nor \ac{TCE} with $\alpha = 2$ converge when $\eta = 0.8$.
Less extreme values of $\alpha$ are able to converge just as well as the \ac{CE} loss while outperforming it.
The \ac{MAE} loss, as expected, failed to converge under the default training parameters.
}
\label{table:resultsNoise100}
\end{table}

\clearpage

\begin{table}[t!]
\centering
\footnotesize
\def\arraystretch{1.2}
\begin{tabular}{ l | c c c || c c c || c c c | }
  & \multicolumn{3}{c||}{MNIST} & \multicolumn{3}{c||}{CIFAR10+} &  \multicolumn{3}{c|}{VSHN} \\
  noise ($\eta$) & 0.0 & 0.4 & 0.8 & 0.0 & 0.4 & 0.8 & 0.0 & 0.4 & 0.8 \\ \hline \hline
  CE  &  99.69 &99.25 &97.77 &92.10 &83.73 &63.59 &96.57 &93.77 &85.50 \\ \hline
  MSE & 99.73 &99.22 &98.01 &\bf{92.15} &84.90 &63.30 &96.95 &94.21 &85.31 \\ \hline
  MAE & - & - &10.28 &10.00 &10.00 &10.00 & - & - & - \\ \hline
  TCE $\alpha=0.5$ & \bf{99.77} &99.30 &98.02 &92.13 &85.19 &63.86 &96.76 &93.94 &85.46\\
  TCE $\alpha=1.0$ & 99.74 &99.33 &98.09 &91.99 &87.12 &63.14 &97.06 &94.56 &86.35\\
  TCE $\alpha=1.5$ & 99.72 &99.46 &98.17 &91.79 &87.99 &65.62 &97.04 &95.51 &87.30\\
  TCE $\alpha=2.0$ & 99.74 &\bf{99.59} &\bf{98.76} &91.82 &\bf{88.40} &\bf{72.95} &\bf{97.09} &\bf{96.15} &\bf{90.44}\\ \hline
\end{tabular}

Top-1 Accuracy (\%)
\vspace{1em}
\caption{
Evaluation of our \ac{TCE} loss against alternatives using a ResNet-20 on MNIST, CIFAR10+, and VSHN.
We used the reference ResNet implementation and default parameters except for the loss function.
A proportion of the training dataset ($\eta$) had its labels replaced randomly. 
Our \ac{TCE} loss offers generally better performance than the \ac{CE} and the \ac{MSE} losses, in particular when the training labels are noisy.
The MNIST dataset is not challenging anymore, and even when training with $80\%$ of noisy labels the default configuration offers excellent performance.
}
\label{table:resultsNoise10}
\end{table}


\subsection{Learning Behavior under Label Noise}

\begin{figure}[t]
\centering
\begin{tikzpicture}
    \begin{axis}[%
    hide axis,xmin=10,xmax=50,ymin=0,ymax=0.4,
	legend columns=4,legend style={cells={align=left}}
    ]
    \addlegendimage{line width=1pt, solid, color=red}
    \addlegendentry{\acs{CE} $\quad$};
    \addlegendimage{line width=1pt, dashed, color=green}
    \addlegendentry{\acs{MSE} $\quad$};
    \addlegendimage{line width=1pt, dash dot, color=blue}
    \addlegendentry{\acs{TCE} $\alpha = 1.0 \quad$};
    \addlegendimage{line width=1pt, dotted, color=orange}
    \addlegendentry{\acs{TCE} $\alpha = 2.0 \quad$};
    \end{axis}
\end{tikzpicture}\\
\subfloat[CIFAR100+, label noise ($\eta$) = 0.0]{
\begin{tikzpicture}
\tikzstyle{every node}=[font=\footnotesize]
\begin{axis}[   height=3.5cm,  width=4.5cm,
scale only axis, ymin=30,ymax=100,xmin=1,xmax=256 ,enlargelimits=false, y label style={at={(axis description cs:-0.1,.5)},anchor=south},  xtick={128,192},ylabel=Top-1 Test Error (\%),  x label style={at={(axis description cs:0.5,-0.1)},anchor=north},  xlabel=Epoch]
\addplot+[solid, color=red, line width=0.5, mark=none, line join=round] table[skip first n=1, x expr=\coordindex+1, y index=1] {results/Tamed_cifar100_nn.CrossEntropyCriterion()_0.0_resnet20/ErrorRate1.log};
\addplot+[dashed, color=green,   line width=0.5, mark=none, line join=round] table[skip first n=1, x expr=\coordindex+1, y index=1]
{results/Tamed_cifar100_nn.ExtendedMSECriterion{prefix=nn.SoftMax(),p=2}_0.0_resnet20/ErrorRate1.log};
\addplot+[dash dot, color=blue,   line width=0.5, mark=none, line join=round] table[skip first n=1, x expr=\coordindex+1, y index=1]
 {results/Tamed_cifar100_nn.ExtendedEntropyCriterion{prefix=nn.LogSoftMax(),p=0}_0.0_resnet20/ErrorRate1.log};
\addplot+[dotted, color=orange,  line width=0.5, mark=none, line join=round] table[skip first n=1, x expr=\coordindex+1, y index=1] 
 {results/Tamed_cifar100_nn.ExtendedEntropyCriterion{prefix=nn.LogSoftMax(),p=-1}_0.0_resnet20/ErrorRate1.log};
\draw[dashed] (100,0) rectangle (224,200);
\end{axis}
\end{tikzpicture}}
~~
\subfloat[CIFAR100+, label noise ($\eta$) = 0.0]{
\begin{tikzpicture}
\tikzstyle{every node}=[font=\footnotesize]
\begin{axis}[   height=3.5cm,  width=4.5cm,
scale only axis, ymin=30,ymax=50,xmin=100,xmax=224 ,enlargelimits=false, y label style={at={(axis description cs:-0.1,.5)},anchor=south}, xtick={128,192}, ylabel=Top-1 Test Error (\%),  x label style={at={(axis description cs:0.5,-0.1)},anchor=north},  xlabel=Epoch]
\addplot+[solid, color=red, line width=0.5, mark=none, line join=round] table[skip first n=1, x expr=\coordindex+1, y index=1] {results/Tamed_cifar100_nn.CrossEntropyCriterion()_0.0_resnet20/ErrorRate1.log};
\addplot+[dashed, color=green,   line width=0.5, mark=none, line join=round] table[skip first n=1, x expr=\coordindex+1, y index=1] 
 {results/Tamed_cifar100_nn.ExtendedMSECriterion{prefix=nn.SoftMax(),p=2}_0.0_resnet20/ErrorRate1.log};
\addplot+[dash dot, color=blue,   line width=0.5, mark=none, line join=round] table[skip first n=1, x expr=\coordindex+1, y index=1] 
{results/Tamed_cifar100_nn.ExtendedEntropyCriterion{prefix=nn.LogSoftMax(),p=0}_0.0_resnet20/ErrorRate1.log};
\addplot+[dotted, color=orange,  line width=0.5, mark=none, line join=round] table[skip first n=1, x expr=\coordindex+1, y index=1]
{results/Tamed_cifar100_nn.ExtendedEntropyCriterion{prefix=nn.LogSoftMax(),p=-1}_0.0_resnet20/ErrorRate1.log};
\end{axis}
\label{fig:detailNoiseless}
\end{tikzpicture}}

\subfloat[CIFAR100+, label noise ($\eta$) = 0.2]{
\begin{tikzpicture}
\tikzstyle{every node}=[font=\footnotesize]
\begin{axis}[   height=3.5cm,  width=4.5cm,
scale only axis, ymin=30,ymax=100,xmin=1,xmax=256 ,enlargelimits=false, y label style={at={(axis description cs:-0.1,.5)},anchor=south},  xtick={128,192},ylabel=Top-1 Test Error (\%),  x label style={at={(axis description cs:0.5,-0.1)},anchor=north},  xlabel=Epoch]
\addplot+[solid, color=red, line width=0.5, mark=none, line join=round] table[skip first n=1, x expr=\coordindex+1, y index=1] {results/Tamed_cifar100_nn.CrossEntropyCriterion()_0.2_resnet20/ErrorRate1.log};
\addplot+[dashed, color=green,   line width=0.5, mark=none, line join=round] table[skip first n=1, x expr=\coordindex+1, y index=1]
{results/Tamed_cifar100_nn.ExtendedMSECriterion{prefix=nn.SoftMax(),p=2}_0.2_resnet20/ErrorRate1.log};
\addplot+[dash dot, color=blue,   line width=0.5, mark=none, line join=round] table[skip first n=1, x expr=\coordindex+1, y index=1]
 {results/Tamed_cifar100_nn.ExtendedEntropyCriterion{prefix=nn.LogSoftMax(),p=0}_0.2_resnet20/ErrorRate1.log};
\addplot+[dotted, color=orange,  line width=0.5, mark=none, line join=round] table[skip first n=1, x expr=\coordindex+1, y index=1] 
 {results/Tamed_cifar100_nn.ExtendedEntropyCriterion{prefix=nn.LogSoftMax(),p=-1}_0.2_resnet20/ErrorRate1.log};
\end{axis}
\end{tikzpicture}}
~~
\subfloat[CIFAR100+, label noise ($\eta$) = 0.4]{
\begin{tikzpicture}
\tikzstyle{every node}=[font=\footnotesize]
\begin{axis}[   height=3.5cm,  width=4.5cm,
scale only axis, ymin=30,ymax=100,xmin=1,xmax=256 ,enlargelimits=false, y label style={at={(axis description cs:-0.1,.5)},anchor=south},  xtick={128,192},ylabel=Top-1 Test Error (\%),  x label style={at={(axis description cs:0.5,-0.1)},anchor=north},  xlabel=Epoch]
\addplot+[solid, color=red, line width=0.5, mark=none, line join=round] table[skip first n=1, x expr=\coordindex+1, y index=1] {results/Tamed_cifar100_nn.CrossEntropyCriterion()_0.4_resnet20/ErrorRate1.log};
\addplot+[dashed, color=green,   line width=0.5, mark=none, line join=round] table[skip first n=1, x expr=\coordindex+1, y index=1] 
 {results/Tamed_cifar100_nn.ExtendedMSECriterion{prefix=nn.SoftMax(),p=2}_0.4_resnet20/ErrorRate1.log};
\addplot+[dash dot, color=blue,   line width=0.5, mark=none, line join=round] table[skip first n=1, x expr=\coordindex+1, y index=1] 
{results/Tamed_cifar100_nn.ExtendedEntropyCriterion{prefix=nn.LogSoftMax(),p=0}_0.4_resnet20/ErrorRate1.log};
\addplot+[dotted, color=orange,  line width=0.5, mark=none, line join=round] table[skip first n=1, x expr=\coordindex+1, y index=1]
{results/Tamed_cifar100_nn.ExtendedEntropyCriterion{prefix=nn.LogSoftMax(),p=-1}_0.4_resnet20/ErrorRate1.log};
\end{axis}
\end{tikzpicture}}

\subfloat[CIFAR100+, label noise ($\eta$) = 0.6]{
\begin{tikzpicture}
\tikzstyle{every node}=[font=\footnotesize]
\begin{axis}[   height=3.5cm,  width=4.5cm,
scale only axis, ymin=30,ymax=100,xmin=1,xmax=256 ,enlargelimits=false, y label style={at={(axis description cs:-0.1,.5)},anchor=south},  xtick={128,192},ylabel=Top-1 Test Error (\%),  x label style={at={(axis description cs:0.5,-0.1)},anchor=north},  xlabel=Epoch]
\addplot+[solid, color=red, line width=0.5, mark=none, line join=round] table[skip first n=1, x expr=\coordindex+1, y index=1] {results/Tamed_cifar100_nn.CrossEntropyCriterion()_0.6_resnet20/ErrorRate1.log};
\addplot+[dashed, color=green,   line width=0.5, mark=none, line join=round] table[skip first n=1, x expr=\coordindex+1, y index=1] 
{results/Tamed_cifar100_nn.ExtendedMSECriterion{prefix=nn.SoftMax(),p=2}_0.6_resnet20/ErrorRate1.log};
\addplot+[dash dot, color=blue,   line width=0.5, mark=none, line join=round] table[skip first n=1, x expr=\coordindex+1, y index=1]
{results/Tamed_cifar100_nn.ExtendedEntropyCriterion{prefix=nn.LogSoftMax(),p=0}_0.6_resnet20/ErrorRate1.log};
\addplot+[dotted, color=orange,  line width=0.5, mark=none, line join=round] table[skip first n=1, x expr=\coordindex+1, y index=1] 
 {results/Tamed_cifar100_nn.ExtendedEntropyCriterion{prefix=nn.LogSoftMax(),p=-1}_0.6_resnet20/ErrorRate1.log};
\end{axis}
\end{tikzpicture}}
~~
\subfloat[CIFAR100+, label noise ($\eta$) = 0.8]{
\begin{tikzpicture}
\tikzstyle{every node}=[font=\footnotesize]
\begin{axis}[   height=3.5cm,  width=4.5cm,
scale only axis, ymin=30,ymax=100,xmin=1,xmax=256 ,enlargelimits=false, y label style={at={(axis description cs:-0.1,.5)},anchor=south}, xtick={128,192}, ylabel=Top-1 Test Error (\%),  x label style={at={(axis description cs:0.5,-0.1)},anchor=north},  xlabel=Epoch]
\addplot+[solid, color=red, line width=0.5, mark=none, line join=round] table[skip first n=1, x expr=\coordindex+1, y index=1] {results/Tamed_cifar100_nn.CrossEntropyCriterion()_0.8_resnet20/ErrorRate1.log};
\addplot+[dashed, color=green,   line width=0.5, mark=none, line join=round] table[skip first n=1, x expr=\coordindex+1, y index=1] 
{results/Tamed_cifar100_nn.ExtendedMSECriterion{prefix=nn.SoftMax(),p=2}_0.8_resnet20/ErrorRate1.log};
\addplot+[dash dot, color=blue,   line width=0.5, mark=none, line join=round] table[skip first n=1, x expr=\coordindex+1, y index=1] 
{results/Tamed_cifar100_nn.ExtendedEntropyCriterion{prefix=nn.LogSoftMax(),p=0}_0.8_resnet20/ErrorRate1.log};
\addplot+[dotted, color=orange,  line width=0.5, mark=none, line join=round] table[skip first n=1, x expr=\coordindex+1, y index=1] 
{results/Tamed_cifar100_nn.ExtendedEntropyCriterion{prefix=nn.LogSoftMax(),p=-1}_0.8_resnet20/ErrorRate1.log};
\end{axis}
\end{tikzpicture}}

\caption{
Test error curves during training ResNet20 on CIFAR100+ under different losses, and different noise ratios. Note that (b) is a detail of (a), where we can observe how the \ac{CE} loss performance worsens even without label noise. The \ac{TCE} and \ac{MSE} losses are more robust to accuracy regressions. 
}
\label{fig:leaningBehavior}
\end{figure}
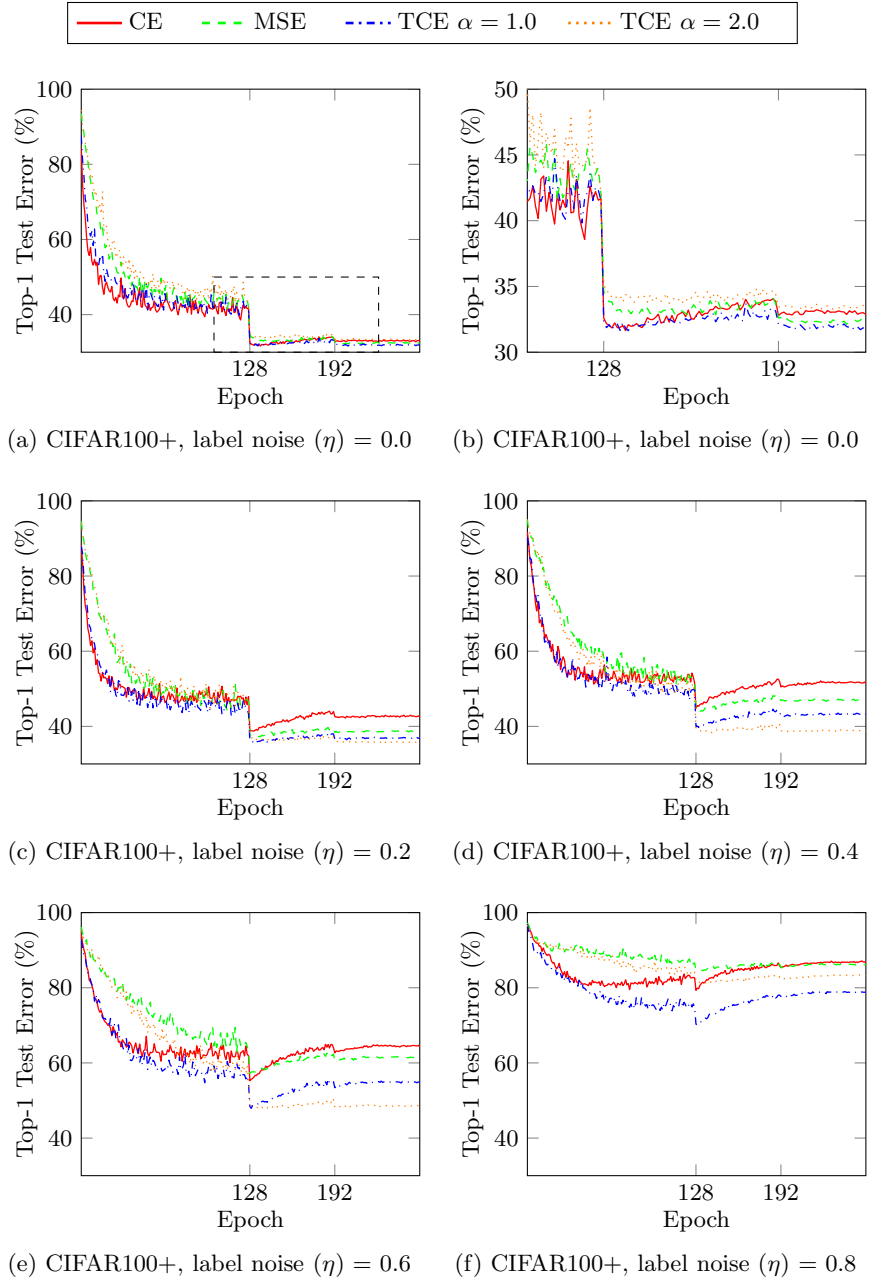

In \figref{leaningBehavior} we show the test set accuracy during training for the CIFAR100+ dataset with different levels of noise.

The first stage of training, with a learning rate of $10^{-1}$, correspond to the annealing stage, and it generally shows little overfitting behavior.
In this stage, the \ac{CE} loss converges the fastest, and the \ac{TCE} with $\alpha=1$ performs similarly. On the other hand, when using $\alpha=2$, the convergence ratio of the \ac{TCE} is similar to that of the \ac{MSE} loss.

At the epoch 128, we reduce the learning rate to $10^{-2}$, and all losses experience a significant drop in the error rate. 
However, after this drop, the networks start to overfit on the mislabeled images, raising the error rate.
Although this effect grows stronger together with the noise ($\eta$), it is still present even with $\eta = 0$, as can be seen in \figref{detailNoiseless}.


\clearpage
\subsection{Top-1 Accuracy vs. Convergence Speed}

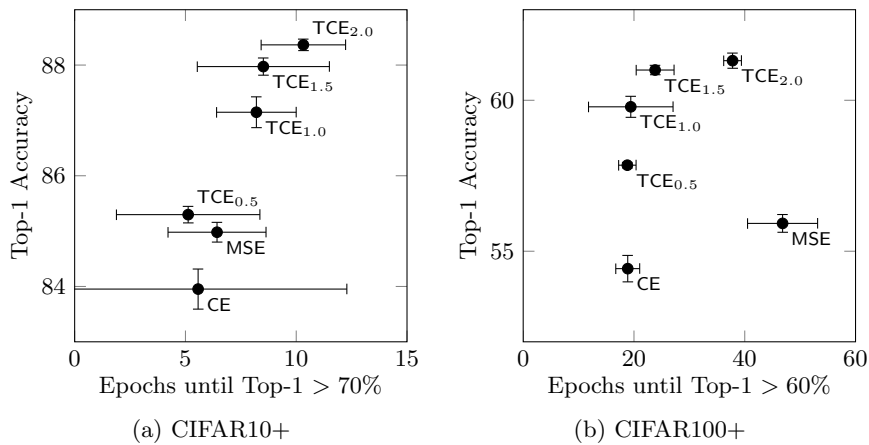
\begin{figure}[t!]
\centering
\subfloat[CIFAR10+]{
\begin{tikzpicture}
\begin{axis}[
x tick label style={font={\footnotesize},yshift=0mm}, 
y tick label style={font={\footnotesize},xshift=0mm},
xlabel style={font={\footnotesize},yshift= 1mm}, 
ylabel style={font={\footnotesize},yshift=-1mm},
height=6cm, width=6cm, xmin=0, xmax=15, ymin=83, ymax=89, 
ylabel=Top-1 Accuracy, xlabel=Epochs until Top-1 $>70\%$]
	\addplot[color=black,mark=*,only marks, point meta=explicit symbolic, nodes near coords,every node near coord/.append style={font={\sffamily\scriptsize},rotate=0, anchor=\myanchor},visualization depends on={value \thisrow{anchor}\as\myanchor},error bars/.cd, y dir=both, y explicit, x dir=both, x explicit] 
	table [x=x,y=y,
	y error plus expr=\thisrow{ysd},
	y error minus expr=\thisrow{ysd},
	x error plus expr=\thisrow{xsd},
	x error minus expr=\thisrow{xsd},
	row sep=\\, meta=Label] {
y ysd x xsd Label anchor\\
83.9525 0.362158 5.57194 6.71303 \acs{CE} {north west}\\
84.98 0.178326 6.42542 2.21151 \acs{MSE} {north west}\\
85.2975 0.149304 5.12303 3.23745 {$\text{TCE}_{0.5}$} {south west}\\
87.1475 0.278373 8.20213 1.79615 {$\text{TCE}_{1.0}$} {north west}\\
87.9725 0.15543 8.51573 2.9806 {$\text{TCE}_{1.5}$} {north west}\\
88.365 0.103763 10.3224 1.90608 {$\text{TCE}_{2.0}$} {south west}\\
	};
\end{axis}
\end{tikzpicture}}
~
\subfloat[CIFAR100+]{
\begin{tikzpicture}
\begin{axis}[
x tick label style={font={\footnotesize},yshift=0mm}, 
y tick label style={font={\footnotesize},xshift=0mm},
xlabel style={font={\footnotesize},yshift= 1mm}, 
ylabel style={font={\footnotesize},yshift=-1mm},
height=6cm, width=6cm, ymin=52, ymax=63, xmin=0, xmax=60, 
ylabel=Top-1 Accuracy, xlabel=Epochs until Top-1 $>60\%$]
	\addplot[color=black,mark=*,only marks, point meta=explicit symbolic, nodes near coords,every node near coord/.append style={font={\sffamily\scriptsize},rotate=0, anchor=\myanchor},visualization depends on={value \thisrow{anchor}\as\myanchor},error bars/.cd, y dir=both, y explicit, x dir=both, x explicit] 
	table [x=x,y=y,
	y error plus expr=\thisrow{ysd},
	y error minus expr=\thisrow{ysd},
	x error plus expr=\thisrow{xsd},
	x error minus expr=\thisrow{xsd},
	row sep=\\, meta=Label] {
y ysd x xsd Label anchor\\
54.4233 0.438072 18.8734 2.15911 \acs{CE} {north west}\\
55.9217 0.294035 46.8294 6.3233 \acs{MSE} {north west}\\
57.85 0.0923038 18.805 1.57347 {$\text{TCE}_{0.5}$} {north west}\\
59.7833 0.348061 19.4241 7.63052 {$\text{TCE}_{1.0}$} {north west}\\
61.0017 0.155874 23.8174 3.41783 {$\text{TCE}_{1.5}$} {north west}\\
61.3117 0.251509 37.7984 1.59971 {$\text{TCE}_{2.0}$} {north west}\\
	};
\end{axis}
\end{tikzpicture}}
\caption{We compare accuracy against convergence speed on CIFAR10+ and CIFAR100+ datasets with $40\%$ of random labels.
The vertical axis shows the accuracy obtained by each loss.
The horizontal axis shows the number of epochs needed to reach the first plateau.
Error bars represent standard deviation.
}
\label{fig:convSpeed}
\end{figure}

In \figref{convSpeed} we evaluate Top-1 Accuracy against convergence speed.
We measure convergence speed by counting how many epochs are necessary for the network to achieve an accuracy threshold. 
We observe a large variance when measuring convergence speed in 10-class datasets like CIFAR10+ (see \figref{convSpeed}a), 
thus our analysis will be based only on the CIFAR100+ results (\figref{convSpeed}b).

For $\alpha \le 1$, the \ac{TCE} loss shows approximately the same convergence speed than the \ac{CE} loss, but better accuracy.
Only for $\alpha > 1$ the \ac{TCE} keeps growing slower, but never reaches the slow convergence speed of the \ac{MSE}.

This results seem to indicate that \ac{TCE} with $\alpha = 1.5$ offers the best trade-off between accuracy and convergence speed.

\section{Conclusions}
We have proposed a new loss function, named \acf{TCE}, that behaves like the popular \ac{CE} loss but is more robust to universal label noise.
The \ac{TCE} loss is friendly to use and can be applied to classification tasks where the \ac{CE} loss is currently used without altering the training parameters.
We found experimentally that the only regularization parameter of the \ac{TCE} loss has a limited effective range $0<\alpha<2$, and changes in $\alpha$ do not have dramatic effects on the performance of the loss.
We expect to extend this work in the future by applying the \ac{TCE} loss to more classification tasks, as well as weakly supervised tasks.
\clearpage

%
%
%
\bibliographystyle{splncs04}

\end{document}